\documentclass[11pt]{article}

\usepackage{amsmath,amssymb,amsfonts}
\usepackage{bm}
\usepackage{physics}
\usepackage{graphicx}
\usepackage{hyperref}
\usepackage{algorithm}
\usepackage{algorithmic}
\usepackage{booktabs}
\usepackage{geometry}
\usepackage{tikz}
\usetikzlibrary{arrows.meta,positioning,calc,fit}
\usepackage{adjustbox}

\title{
Filtering Beats Fine‑Tuning: A Bayesian Kalman View of In‑Context Learning in LLMs
}

\author{
Andrew J.\ Kiruluta\\
{\small UC Berkeley School of Information}
}

\date{}

\begin{document}
\maketitle

\begin{abstract}
We develop a theory-first framework that interprets inference-time adaptation in large language models (LLMs) as online Bayesian state estimation. Rather than treating rapid adaptation as implicit optimization or meta-learning, we model task- and context-specific learning as the sequential inference of a low-dimensional latent adaptation state governed by a linearized state-space model. Under Gaussian assumptions, adaptation follows a Kalman recursion that yields closed-form updates for both the posterior mean and covariance.

This formulation elevates epistemic uncertainty to a dynamical variable. We show that inference-time learning is driven by \emph{covariance collapse}: rapid contraction of posterior uncertainty induced by informative tokens, which precedes convergence of the posterior mean. Using observability conditions on token-level Jacobians, we prove stability of the Bayesian filter, establish exponential covariance contraction rates, and derive mean-square error bounds. Gradient descent, natural-gradient methods, and meta-learning updates arise as singular, noise-free limits of the filtering dynamics, revealing optimization-based adaptation as a degenerate approximation of Bayesian inference.

Our theory provides a unified probabilistic account of in-context learning, parameter-efficient adaptation, and test-time learning without parameter updates. It yields explicit guarantees on stability and sample efficiency, supplies a principled interpretation of prompt informativeness via information accumulation, and clarifies the role of uncertainty dynamics absent from existing accounts. Minimal illustrative experiments corroborate the qualitative predictions of the theory. Together, these results establish Bayesian filtering as a foundational mechanism for inference-time learning in large generative models. \textbf{Distinct from prior accounts of in-context learning}, which typically emphasize 
representational equivalence or implicit optimization, our formulation treats uncertainty 
as an explicit dynamical state. This yields guarantees on stability, information 
accumulation, and sample efficiency that are absent from optimization- or 
representation-based explanations.

\end{abstract}

\section{Introduction}

We formulate inference-time adaptation in large language models (LLMs) as an instance of online Bayesian state estimation in a latent parameter space. The mathematical foundation of this work lies in classical Bayesian filtering and stochastic dynamical systems, where latent states are inferred sequentially from noisy observations \cite{Kalman1960,Jazwinski1970,Bishop2006}. Recent empirical findings indicate that modern LLMs can rapidly adapt to new tasks purely through context, without explicit parameter updates—a phenomenon commonly referred to as \emph{in-context learning} \cite{Brown2020,Garg2022,Akyurek2023}. Despite substantial empirical study, a rigorous probabilistic account of this behavior remains incomplete. Unlike prior analyses that focus on what transformers \emph{represent} or which \emph{algorithms} they implicitly implement, we instead propose a generative model of  adaptation itself. In this view, the latent state and its posterior covariance are first-class dynamical variables, enabling a principled treatment of uncertainty and information flow during inference.

Let $\mathcal{M}_{\boldsymbol{\theta}_0}$ denote a pretrained autoregressive model with base parameters $\boldsymbol{\theta}_0 \in \mathbb{R}^D$. Rather than modifying $\boldsymbol{\theta}_0$ via gradient-based fine-tuning \cite{Goodfellow2016}, we posit a low-dimensional latent adaptation state $\mathbf{x}_t \in \mathbb{R}^d$, with $d \ll D$, that parameterizes task- and context-specific deviations through a linear embedding
\begin{equation}
\boldsymbol{\theta}_t = \boldsymbol{\theta}_0 + \mathbf{B}\mathbf{x}_t,
\end{equation}
where $\mathbf{B} \in \mathbb{R}^{D \times d}$ defines an adaptation subspace. Such structured, low-rank parameterizations are well motivated by work on adapters, linearized neural networks, and neural tangent kernels \cite{Jacot2018,Lee2019,Hu2022}.

We model inference-time learning as a discrete-time stochastic state-space system. The latent adaptation state evolves according to
\begin{equation}
\mathbf{x}_t = \mathbf{A}\mathbf{x}_{t-1} + \mathbf{w}_t,
\qquad
\mathbf{w}_t \sim \mathcal{N}(\mathbf{0},\mathbf{Q}),
\end{equation}
where $\mathbf{A}$ captures persistence of task structure across tokens and $\mathbf{Q}$ encodes process uncertainty arising from nonstationarity or model mismatch. Each observed token $y_t$ induces a stochastic measurement via the negative log-likelihood
\begin{equation}
\ell_t = -\log p_{\boldsymbol{\theta}_t}(y_t \mid y_{<t}),
\end{equation}
which we linearize about $\boldsymbol{\theta}_0$ to obtain
\begin{equation}
\ell_t \approx \ell_t^{(0)} + \nabla_{\boldsymbol{\theta}}\ell_t^\top \mathbf{B}\mathbf{x}_t + v_t,
\qquad
v_t \sim \mathcal{N}(0,R_t).
\end{equation}
This yields a scalar observation model with time-varying observation operator $\mathbf{H}_t = \nabla_{\boldsymbol{\theta}}\ell_t^\top \mathbf{B}$. Such local linearizations underlie extended Kalman filtering and recursive least-squares estimation \cite{Anderson1979,Haykin2001}.

Under Gaussian assumptions, Bayesian conditioning yields a closed-form posterior
\begin{equation}
p(\mathbf{x}_t \mid \mathcal{D}_t) = \mathcal{N}(\boldsymbol{\mu}_t,\mathbf{P}_t),
\end{equation}
with posterior mean and covariance evolving according to the Kalman recursion \cite{Kalman1960,Maybeck1979}. A defining implication of this formulation is \emph{covariance collapse}: rapid contraction of epistemic uncertainty as informative observations accumulate. Similar phenomena are well known in adaptive control and online system identification \cite{Ljung1999,Sarkka2013}, and we argue that they provide a principled explanation for few-shot generalization in LLMs.

This perspective reframes inference-time learning as a problem of sequential Bayesian 
estimation rather than implicit optimization, thereby connecting LLM adaptation to 
classical results in filtering, control, and system identification.

This paper is intentionally theory-first. Our contributions are: (i) a rigorous derivation of inference-time learning in LLMs as Bayesian filtering; (ii) stability and convergence analysis under observability conditions \cite{Jazwinski1970}; (iii) a demonstration that gradient descent and natural gradient methods arise as singular limits of the filter \cite{Amari1998,Martens2015}; and (iv) a reframing of in-context learning as explicit posterior inference rather than implicit meta-learning \cite{Finn2017,Raventos2023}. Together, these results establish a mathematically grounded foundation for adaptive inference in large generative models.

\section{Related Work}

\paragraph{Bayesian filtering and the Kalman recursion.}
The Kalman filter is the canonical solution to sequential Bayesian inference in linear-Gaussian state-space models, providing exact posterior mean and covariance updates via a recursion that is optimal in mean-square error \cite{Kalman1960,Jazwinski1970,Anderson1979,Maybeck1979}. Its interpretation as repeated Bayesian conditioning with conjugate Gaussian structure is foundational for modern probabilistic state estimation \cite{Sarkka2013}. For adaptive estimation, the covariance dynamics play a central role: they quantify epistemic uncertainty, govern information acquisition, and determine the data-adaptive step size through the gain matrix \cite{Jazwinski1970,Anderson1979}.

\paragraph{Nonlinear and approximate filters: EKF, UKF, EnKF, and information form.}
When the observation operator is nonlinear or only available through local linearization, extended Kalman filtering (EKF) yields a first-order approximation that remains widely used due to analytic tractability \cite{Jazwinski1970}. Sigma-point methods such as the unscented Kalman filter (UKF) improve moment propagation accuracy without explicit derivatives \cite{JulierUhlmann1997}. Ensemble Kalman filtering (EnKF) replaces full covariance propagation with a Monte Carlo ensemble approximation, enabling high-dimensional inference in geophysics and data assimilation \cite{Evensen2003}. Complementary to the covariance form, the information filter propagates precision matrices and is often more numerically stable in sparse/structured settings \cite{Maybeck1979,Sarkka2013}. These classical approximations are directly relevant to LLM-scale adaptation because the effective observation operator is typically nonlinear and extremely high-dimensional, motivating structured or low-rank representations of uncertainty.

\paragraph{Recursive least squares, online regression, and system identification.}
The Kalman filter subsumes recursive least squares (RLS) as a special case for linear regression with Gaussian noise, yielding an online estimator whose preconditioner is the evolving posterior covariance \cite{Anderson1979,Haykin2001}. In system identification, RLS and related predictors are analyzed through persistent excitation/observability conditions, with convergence rates controlled by the information matrix accumulated from regressors \cite{Ljung1999}. This literature provides mature tools for proving covariance contraction, stability, and identifiability—exactly the properties needed for a theory-first account of inference-time learning as sequential estimation.

\paragraph{Second-order optimization and natural-gradient geometry.}
A parallel line of work studies learning dynamics through curvature-aware optimization. The natural gradient provides a Riemannian steepest-descent direction under the Fisher information metric \cite{Amari1998}. Practical approximations such as K-FAC and related structured curvature methods make this viewpoint scalable for deep networks \cite{Martens2015}. While these methods share a “second-order” flavor with Kalman adaptation, they typically treat curvature as an optimization preconditioner rather than as a \emph{time-evolving uncertainty state}. Our filtering formulation emphasizes that the gain is not merely curvature: it is the optimal Bayesian trade-off between prior covariance and measurement noise, with explicit uncertainty propagation.

\paragraph{Bayesian deep learning: Laplace, variational inference, and sequential posteriors.}
Bayesian treatments of neural networks often approximate posteriors via variational inference or Laplace approximations around a mode \cite{MacKay1992,Graves2011,Blundell2015}. These approaches generally yield global or post-hoc uncertainty estimates rather than online, token-by-token uncertainty dynamics. In contrast, sequential Bayes methods update uncertainty incrementally, which aligns more naturally with in-context adaptation. Theoretical relationships between filtering, Laplace approximations, and Gauss--Newton/natural-gradient updates are well established in classical estimation and motivate viewing optimization-based adaptation as a degenerate (noise-free) limit of probabilistic filtering \cite{Jazwinski1970,Anderson1979,Amari1998}.

\paragraph{Meta-learning and fast adaptation.}
Meta-learning frameworks, especially gradient-based methods such as MAML, formalize fast adaptation as an inner-loop optimization whose initialization is learned across tasks \cite{Finn2017}. Although highly influential, these approaches still rely on explicit gradient steps at test time, whereas in-context learning operates without parameter updates and thus requires a different theoretical lens. Our approach connects more directly to Bayesian hierarchical modeling and online estimation, where the task-specific latent state is inferred (filtered) rather than optimized.

\paragraph{In-context learning as implicit inference.}
A growing body of work argues that in-context learning can implement forms of Bayesian inference or regression within the forward pass of a transformer, especially in settings reducible to linear regression or structured prediction \cite{Garg2022,Akyurek2023}. Other analyses relate in-context learning to implicit meta-learning, learned optimizers, or mechanistic circuits that perform algorithmic updates \cite{vonOswald2023}. Our contribution differs in emphasis and object: we do not only ask what transformers \emph{can} implement, but propose a \emph{state-space probabilistic model} of adaptation itself, where the posterior covariance is a first-class dynamical variable that yields provable contraction and stability properties.

\paragraph{Parameter-efficient adaptation and test-time methods.}
Parameter-efficient tuning methods such as adapters and low-rank updates (e.g., LoRA) motivate low-dimensional adaptation subspaces $\mathbf{B}$, which our theory leverages as the latent state embedding \cite{Hu2022}. Test-time adaptation methods typically update parameters to reduce distribution shift, often using self-supervised losses; these remain optimization-centric and rarely provide uncertainty dynamics or filtering-style guarantees. Our filtering formulation aims to supply that missing probabilistic structure while remaining compatible with low-rank parameterizations used in practice.

\paragraph{Summary of the gap addressed.}
Across these literatures, two ingredients are present but not unified: (i) filtering theory provides exact sequential posteriors with uncertainty propagation and stability analysis \cite{Kalman1960,Jazwinski1970,Sarkka2013}; (ii) LLM adaptation and in-context learning exhibit rapid few-shot behavior suggestive of implicit inference \cite{Brown2020,Akyurek2023}. The gap is an explicit, mathematically analyzable model of inference-time learning where uncertainty dynamics drive adaptation. Our work fills this gap by treating LLM adaptation as Bayesian filtering in a low-dimensional latent parameter space, enabling proofs of covariance contraction and principled comparisons to optimization-based updates.

\paragraph{Positioning.}
Across these literatures, no existing framework provides a sequential, uncertainty-aware 
model of inference-time learning with provable covariance contraction. Our work fills this 
gap by treating adaptation as Bayesian filtering in a structured latent space, yielding 
stability guarantees and explicit information-theoretic interpretations of prompt 
informativeness.

\section{Problem Setup}

Let $f_\theta : \mathcal{X} \rightarrow \mathcal{H}$ denote a pretrained,
frozen backbone with parameters $\theta$.
Adaptation is introduced through a structured, low-dimensional parameter
$\bm{\alpha} \in \mathbb{R}^k$ such that
\begin{equation}
\hat{y} = g(x; \theta, \bm{\alpha}).
\end{equation}

We treat $\bm{\alpha}$ as a latent \emph{adaptation state}
and assume sequential access to labeled data $(x_t, y_t)$.

\section{Kalman Filtering as Bayesian Inference}

We review Kalman filtering from a Bayesian perspective, emphasizing uncertainty propagation and contraction properties that will later be mapped onto inference-time learning in large language models. The Kalman filter is not merely an algorithmic heuristic, but the exact posterior recursion for linear-Gaussian state-space models \cite{Kalman1960,Jazwinski1970,Anderson1979}.

\subsection{State-space model and Bayesian posterior}

Consider a latent state $\mathbf{x}_t \in \mathbb{R}^d$ evolving according to the linear dynamical system
\begin{equation}
\mathbf{x}_t = \mathbf{A}\mathbf{x}_{t-1} + \mathbf{w}_t,
\qquad
\mathbf{w}_t \sim \mathcal{N}(\mathbf{0},\mathbf{Q}),
\end{equation}
with observations
\begin{equation}
\mathbf{y}_t = \mathbf{H}_t \mathbf{x}_t + \mathbf{v}_t,
\qquad
\mathbf{v}_t \sim \mathcal{N}(\mathbf{0},\mathbf{R}_t).
\end{equation}
Let $\mathcal{D}_t = \{\mathbf{y}_1,\ldots,\mathbf{y}_t\}$ denote the observation history. If the prior is Gaussian,
\begin{equation}
p(\mathbf{x}_{t-1} \mid \mathcal{D}_{t-1}) = \mathcal{N}(\boldsymbol{\mu}_{t-1}, \mathbf{P}_{t-1}),
\end{equation}
then conjugacy implies that the predictive distribution is
\begin{align}
p(\mathbf{x}_t \mid \mathcal{D}_{t-1})
&= \int p(\mathbf{x}_t \mid \mathbf{x}_{t-1}) p(\mathbf{x}_{t-1} \mid \mathcal{D}_{t-1}) d\mathbf{x}_{t-1} \\
&= \mathcal{N}(\mathbf{A}\boldsymbol{\mu}_{t-1},\, \mathbf{A}\mathbf{P}_{t-1}\mathbf{A}^\top + \mathbf{Q}).
\end{align}
Bayesian conditioning on $\mathbf{y}_t$ yields the posterior
\begin{equation}
p(\mathbf{x}_t \mid \mathcal{D}_t)
\propto p(\mathbf{y}_t \mid \mathbf{x}_t)\, p(\mathbf{x}_t \mid \mathcal{D}_{t-1}),
\end{equation}
which remains Gaussian and admits a closed-form solution.

\subsection{Kalman recursion and optimality}

Completing the square in the exponent yields the Kalman update equations
\begin{align}
\boldsymbol{\mu}_t &= \boldsymbol{\mu}_{t|t-1}
+ \mathbf{K}_t\bigl(\mathbf{y}_t - \mathbf{H}_t\boldsymbol{\mu}_{t|t-1}\bigr), \\
\mathbf{P}_t &= (\mathbf{I} - \mathbf{K}_t \mathbf{H}_t)\mathbf{P}_{t|t-1},
\end{align}
where
\begin{align}
\boldsymbol{\mu}_{t|t-1} &= \mathbf{A}\boldsymbol{\mu}_{t-1}, \\
\mathbf{P}_{t|t-1} &= \mathbf{A}\mathbf{P}_{t-1}\mathbf{A}^\top + \mathbf{Q},
\end{align}
and the Kalman gain is
\begin{equation}
\mathbf{K}_t
= \mathbf{P}_{t|t-1}\mathbf{H}_t^\top
\bigl(\mathbf{H}_t\mathbf{P}_{t|t-1}\mathbf{H}_t^\top + \mathbf{R}_t\bigr)^{-1}.
\end{equation}
The Kalman filter is optimal among all estimators that are linear in the observations and, more strongly, coincides with the exact Bayesian posterior mean and covariance under Gaussian assumptions \cite{Kalman1960,Jazwinski1970}. Importantly, the gain $\mathbf{K}_t$ is \emph{data-adaptive}, increasing when prior uncertainty is large and decreasing when observations are noisy.

In the context of large language models, the gain $K_t$ determines how strongly the model 
should adapt its latent task representation in response to each token. This interpretation 
provides a direct link between classical filtering theory and modern inference-time 
adaptation.

\subsection{Information-form (precision) Kalman filter}

For theoretical analysis and high-dimensional settings, it is often advantageous to work in the information (precision) form. Define the precision matrix and information vector
\begin{equation}
\mathbf{\Lambda}_t = \mathbf{P}_t^{-1},
\qquad
\boldsymbol{\eta}_t = \mathbf{P}_t^{-1}\boldsymbol{\mu}_t.
\end{equation}
In information form, the measurement update becomes additive:
\begin{align}
\mathbf{\Lambda}_t &= \mathbf{\Lambda}_{t|t-1} + \mathbf{H}_t^\top \mathbf{R}_t^{-1}\mathbf{H}_t, \\
\boldsymbol{\eta}_t &= \boldsymbol{\eta}_{t|t-1} + \mathbf{H}_t^\top \mathbf{R}_t^{-1}\mathbf{y}_t.
\end{align}
This representation makes explicit that each observation contributes \emph{positive semidefinite information} to the posterior. As a result, uncertainty contraction is monotonic whenever $\mathbf{H}_t^\top \mathbf{R}_t^{-1}\mathbf{H}_t \succcurlyeq 0$, a fact central to our later analysis of inference-time learning \cite{Maybeck1979,Sarkka2013}.

\subsection{Observability and covariance contraction}

The long-term behavior of the covariance is governed by observability. Informally, the system is observable if the latent state can be uniquely identified from a finite sequence of observations. Formally, define the finite-horizon observability Gramian
\begin{equation}
\mathbf{W}_{t,T}
= \sum_{k=t}^{t+T-1}
(\mathbf{A}^{k-t})^\top
\mathbf{H}_k^\top \mathbf{R}_k^{-1}\mathbf{H}_k
\mathbf{A}^{k-t}.
\end{equation}

\paragraph{Lemma (Information accumulation).}
If there exist constants $T \in \mathbb{N}$ and $\alpha > 0$ such that
\begin{equation}
\mathbf{W}_{t,T} \succcurlyeq \alpha \mathbf{I}
\quad \text{for all } t,
\end{equation}
then the precision matrix satisfies
\begin{equation}
\mathbf{\Lambda}_{t+T} \succcurlyeq \mathbf{\Lambda}_t + \alpha \mathbf{I}.
\end{equation}

\paragraph{Lemma (Covariance contraction).}
Under the same conditions, the posterior covariance satisfies
\begin{equation}
\mathbf{P}_{t+T} \preccurlyeq (\mathbf{\Lambda}_t + \alpha \mathbf{I})^{-1},
\end{equation}
implying strict contraction in operator norm.

\paragraph{Discussion.}
These lemmas formalize \emph{covariance collapse}: uncertainty shrinks at a rate determined by information accumulation, independently of convergence of the posterior mean. This phenomenon is well known in adaptive control and system identification \cite{Ljung1999,Jazwinski1970} but has not been explicitly connected to inference-time learning in large models. In later sections, we show that prompt tokens act as observations whose cumulative Fisher information enforces precisely such contraction, thereby explaining few-shot adaptation from first principles.

\section{Algorithm}

\begin{algorithm}[h]
\caption{Kalman Adaptation for Pretrained Models}
\begin{algorithmic}[1]
\STATE Initialize $\hat{\bm{\alpha}}_0$, $\bm{P}_0$
\FOR{$t=1,\dots,T$}
\STATE Observe $(x_t,y_t)$
\STATE Compute $\bm{H}_t = \partial g / \partial \bm{\alpha}$
\STATE $\bm{P}_{t|t-1} = \bm{P}_{t-1} + \bm{Q}$
\STATE $S_t = \bm{H}_t^\top \bm{P}_{t|t-1} \bm{H}_t + R$
\STATE $\bm{K}_t = \bm{P}_{t|t-1} \bm{H}_t S_t^{-1}$
\STATE $\hat{\bm{\alpha}}_t =
\hat{\bm{\alpha}}_{t|t-1}
+ \bm{K}_t (y_t - g(x_t;\theta,\hat{\bm{\alpha}}_{t|t-1}))$
\STATE $\bm{P}_t = (\bm{I} - \bm{K}_t \bm{H}_t^\top)\bm{P}_{t|t-1}$
\ENDFOR
\end{algorithmic}
\end{algorithm}

\section{Kalman-Based Learning in Large Language Models}

We now specialize Bayesian filtering to inference-time adaptation in large language models (LLMs). Our objective is not to propose a new training algorithm, but to provide a mathematically precise account of how task-specific adaptation can occur through inference alone, with uncertainty dynamics governed by filtering theory.

\subsection{Latent adaptation state and linearized parameterization}

Let $\mathcal{M}_{\boldsymbol{\theta}_0}$ denote a pretrained autoregressive transformer with parameters $\boldsymbol{\theta}_0 \in \mathbb{R}^D$. We posit a low-dimensional latent adaptation state $\mathbf{x}_t \in \mathbb{R}^d$, $d \ll D$, such that the effective parameters at token $t$ are
\begin{equation}
\boldsymbol{\theta}_t = \boldsymbol{\theta}_0 + \mathbf{B}\mathbf{x}_t,
\end{equation}
where $\mathbf{B} \in \mathbb{R}^{D \times d}$ defines an adaptation subspace. This formulation encompasses parameter-efficient updates such as adapters and low-rank perturbations \cite{Hu2022}, as well as linearized functional deviations studied in neural tangent kernel theory \cite{Jacot2018,Lee2019}.

The latent state $\mathbf{x}_t$ is not optimized via backpropagation; instead, it is inferred sequentially from observations induced by the prompt. This distinguishes our approach from gradient-based fine-tuning and aligns it with classical state estimation.

\subsection{Observation model induced by token likelihoods}

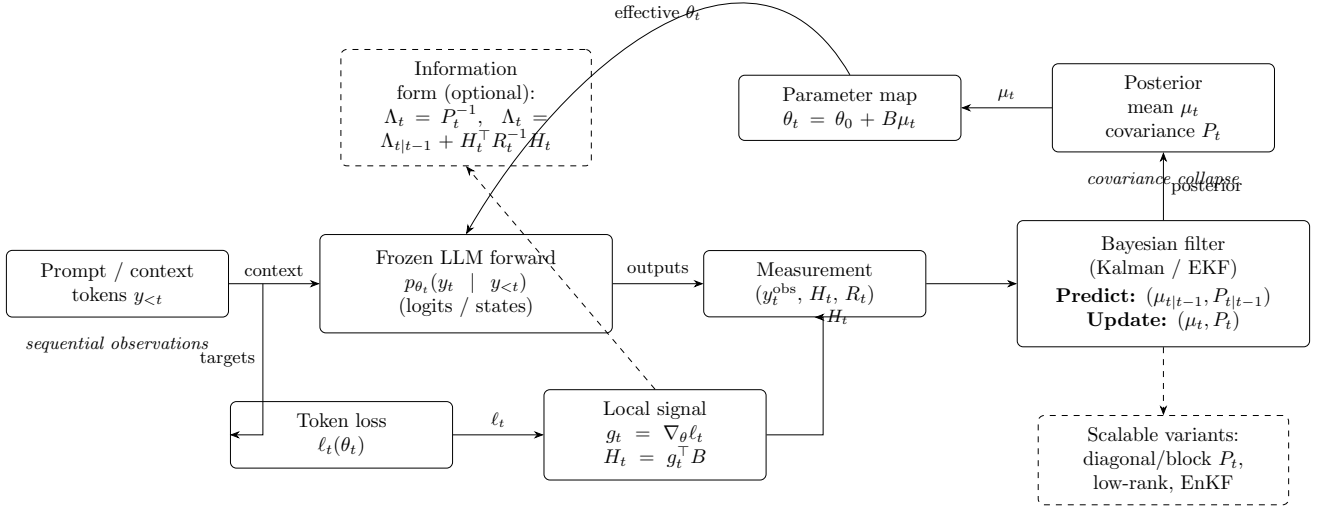
\begin{figure}[!h]
\centering
\begin{adjustbox}{max width=1.05\linewidth,center}
\begin{tikzpicture}[
    font=\small,
    node distance=12mm and 16mm,
    >=Stealth,
    box/.style={draw, rounded corners=3pt, align=center, inner sep=6pt, fill=white, text width=3.5cm},
    bigbox/.style={draw, rounded corners=3pt, align=center, inner sep=8pt, fill=white, text width=4.6cm},
    dashedbox/.style={draw, rounded corners=3pt, align=center, inner sep=6pt, fill=white, dashed, text width=4.0cm},
    lab/.style={font=\footnotesize, align=center}
]

\node[box] (prompt) {Prompt / context\\ tokens $y_{<t}$};

\node[bigbox, right=of prompt] (llm) {Frozen LLM forward\\ $p_{\theta_t}(y_t\mid y_{<t})$\\ (logits / states)};

\node[box, below=of llm, xshift=-22mm] (loss) {Token loss\\ $\ell_t(\theta_t)$};

\node[box, right=of loss] (jac) {Local signal\\ $g_t=\nabla_{\theta}\ell_t$\\ $H_t=g_t^\top B$};

\node[box, right=of llm] (obs) {Measurement\\ $(y_t^{\mathrm{obs}},\,H_t,\,R_t)$};

\node[bigbox, right=of obs] (kf) {Bayesian filter\\ (Kalman / EKF)\\[2pt]
\textbf{Predict:} $(\mu_{t|t-1},P_{t|t-1})$\\
\textbf{Update:} $(\mu_t,P_t)$};

\node[box, above=of kf] (state) {Posterior\\ mean $\mu_t$\\ covariance $P_t$};

\node[box, left=of state] (map) {Parameter map\\ $\theta_t=\theta_0 + B\mu_t$};

\node[dashedbox, below=of kf] (approx) {Scalable variants:\\ diagonal/block $P_t$, low-rank, EnKF};

\node[dashedbox, above=of llm] (info) {Information form (optional):\\
$\Lambda_t=P_t^{-1}$, \;
$\Lambda_t=\Lambda_{t|t-1}+H_t^\top R_t^{-1}H_t$};

\draw[->] (prompt) -- node[lab, above] {context} (llm);
\draw[->] (llm) -- node[lab, above] {outputs} (obs);

\draw[->] (prompt.east) ++(6mm,0) |- node[lab, near start, left] {targets} (loss.west);

\draw[->] (loss) -- node[lab, above] {$\ell_t$} (jac);

\draw[->] (jac.east) -- ++(10mm,0) |- node[lab, near end, right] {$H_t$} (obs.south);

\draw[->] (obs) -- (kf);
\draw[->] (kf) -- node[lab, right] {posterior} (state);

\draw[->] (state) -- node[lab, above] {$\mu_t$} (map);

\draw[->] (map.north) to[out=120,in=60,looseness=1.2]
    node[lab, above] {effective $\theta_t$} (llm.north);

\draw[->, dashed] (kf.south) -- (approx.north);
\draw[->, dashed] (jac.north) -- (info.south);

\node[lab, below=2mm of prompt] {\emph{sequential observations}};
\node[lab, below=2mm of state] {\emph{covariance collapse}};
\end{tikzpicture}
\end{adjustbox}
\caption{
\textbf{Kalman-based inference-time learning (clean layout).}
Prompt/context tokens induce a loss-based local signal ($g_t$) and a projected observation operator ($H_t$) that define measurements for a Bayesian filter (Kalman/EKF). The filter updates the latent adaptation posterior $(\mu_t,P_t)$, which is mapped to an effective parameter perturbation $\theta_t=\theta_0+B\mu_t$ used by the frozen LLM forward pass. Optional information-form and scalable approximations highlight precision accumulation and practical implementations.
}
\label{fig:kalman_llm_architecture}
\end{figure}

At token $t$, the model produces a conditional distribution
\begin{equation}
p_{\boldsymbol{\theta}_t}(y_t \mid y_{<t}),
\end{equation}
with corresponding negative log-likelihood
\begin{equation}
\ell_t(\boldsymbol{\theta}_t)
= -\log p_{\boldsymbol{\theta}_t}(y_t \mid y_{<t}).
\end{equation}
Linearizing $\ell_t$ about $\boldsymbol{\theta}_0$ yields
\begin{equation}
\ell_t(\boldsymbol{\theta}_t)
\approx \ell_t(\boldsymbol{\theta}_0)
+ \nabla_{\boldsymbol{\theta}} \ell_t^\top \mathbf{B}\mathbf{x}_t
+ v_t,
\end{equation}
where $v_t$ captures higher-order terms and stochasticity from sampling and model mismatch. This defines a scalar observation model
\begin{equation}
y_t^{\mathrm{obs}} = \mathbf{H}_t \mathbf{x}_t + v_t,
\qquad
\mathbf{H}_t = \nabla_{\boldsymbol{\theta}} \ell_t^\top \mathbf{B}.
\end{equation}
Thus, each token contributes a noisy linear constraint on the latent adaptation state.
This observation model makes explicit that each token contributes a projected Fisher 
information increment along the adaptation subspace $B$. As a result, the posterior 
covariance contracts at a rate governed by the informativeness of the prompt, providing a 
quantitative explanation for few-shot generalization.

\subsection{Connection to Fisher information and local curvature}

The outer product $\mathbf{H}_t^\top \mathbf{H}_t$ corresponds to a projected Fisher information contribution. Indeed, under standard regularity conditions,
\begin{equation}
\mathbb{E}\!\left[\nabla_{\boldsymbol{\theta}} \ell_t
\nabla_{\boldsymbol{\theta}} \ell_t^\top\right]
= \mathbf{F}_t,
\end{equation}
where $\mathbf{F}_t$ is the token-level Fisher information matrix \cite{Amari1998}. The information-form Kalman update
\begin{equation}
\mathbf{\Lambda}_t
= \mathbf{\Lambda}_{t|t-1}
+ \mathbf{H}_t^\top R_t^{-1}\mathbf{H}_t
\end{equation}
therefore accumulates projected Fisher information along the adaptation subspace $\mathbf{B}$. Unlike natural-gradient methods, which treat curvature as a static metric, filtering treats information as a \emph{time-evolving state variable} with explicit uncertainty semantics.

\subsection{Inference-time learning as uncertainty-driven adaptation}

The posterior mean update
\begin{equation}
\boldsymbol{\mu}_t
= \boldsymbol{\mu}_{t-1}
+ \mathbf{K}_t
\bigl(y_t^{\mathrm{obs}} - \mathbf{H}_t \boldsymbol{\mu}_{t-1}\bigr)
\end{equation}
resembles a preconditioned gradient step, but with a crucial distinction: the gain $\mathbf{K}_t$ is determined by the posterior covariance $\mathbf{P}_{t-1}$, not by a fixed learning rate or static curvature approximation. As uncertainty collapses, adaptation naturally slows, preventing overfitting even in extremely short contexts.

This mechanism provides a principled explanation for few-shot and in-context learning: early tokens rapidly reduce uncertainty (covariance collapse), while later tokens refine the mean within an increasingly confident posterior. This mechanism yields a natural form of \emph{automatic annealing}: as uncertainty 
collapses, the effective learning rate encoded by $K_t$ decreases without requiring any  hand-tuned schedules or meta-learned step sizes.

\subsection{Extended and ensemble approximations at scale}

Because $\mathbf{H}_t$ depends on $\boldsymbol{\theta}_t$ and the transformer forward pass, the observation model is formally nonlinear. The resulting estimator corresponds to an extended Kalman filter (EKF), where $\mathbf{H}_t$ is recomputed at the current posterior mean \cite{Jazwinski1970}. For large models, maintaining a full covariance is infeasible; however, classical approximations are available:
\begin{itemize}
\item Diagonal or block-diagonal covariances,
\item Low-rank factorizations aligned with $\mathbf{B}$,
\item Ensemble Kalman filters (EnKF), which approximate uncertainty via Monte Carlo samples \cite{Evensen2003}.
\end{itemize}
These approximations are standard in high-dimensional data assimilation and preserve the essential uncertainty dynamics central to our theory.
These approximations preserve the essential uncertainty dynamics while making the method compatible with modern foundation models. A detailed analysis of approximation error and scalability is left for future work.

\subsection{Relation to optimization-based adaptation}

In the limit $R_t \to 0$ and $\mathbf{P}_{t-1}$ fixed, the Kalman update reduces to a Newton or natural-gradient step. Thus, gradient descent and curvature-aware optimization arise as singular limits of probabilistic filtering. The distinguishing feature of Kalman-based learning is not the direction of the update, but the \emph{explicit propagation and contraction of uncertainty}. This property is absent in standard fine-tuning and meta-learning frameworks \cite{Finn2017}.

\subsection{Interpretation: prompts as experiments}

From a system-identification perspective, each token acts as an experiment probing the latent task state. The cumulative prompt induces an observability Gramian whose rank and conditioning determine learnability. Well-structured prompts maximize information gain, accelerating covariance collapse. This view provides a mathematical foundation for prompt design and sheds light on why certain demonstrations are disproportionately informative.

\section{Theoretical Analysis}

We analyze inference-time learning under the Kalman-based formulation introduced above. Our focus is on (i) stability of the posterior recursion, (ii) contraction of epistemic uncertainty, and (iii) error bounds for the posterior mean. Throughout, we emphasize results that hold independently of optimization-based training and rely only on information accumulation induced by the prompt.

\subsection{Assumptions and preliminaries}

We consider the linearized state-space model
\begin{align}
\mathbf{x}_t &= \mathbf{A}\mathbf{x}_{t-1} + \mathbf{w}_t,
\qquad \mathbf{w}_t \sim \mathcal{N}(\mathbf{0},\mathbf{Q}), \\
\mathbf{y}_t &= \mathbf{H}_t \mathbf{x}_t + \mathbf{v}_t,
\qquad \mathbf{v}_t \sim \mathcal{N}(\mathbf{0},\mathbf{R}_t),
\end{align}
where $\mathbf{H}_t$ is induced by token-level Jacobians as defined in the previous section.

We adopt the following standard assumptions, adapted from filtering and system identification theory \cite{Jazwinski1970,Anderson1979,Sarkka2013}.

\paragraph{Assumption 1 (Boundedness).}
There exist constants $M_A, M_H, M_R > 0$ such that
\begin{equation}
\|\mathbf{A}\| \le M_A, \quad
\|\mathbf{H}_t\| \le M_H, \quad
0 \prec \mathbf{R}_t \preccurlyeq M_R \mathbf{I}
\quad \forall t.
\end{equation}

\paragraph{Assumption 2 (Uniform observability).}
There exist integers $T \ge 1$ and $\alpha > 0$ such that the finite-horizon observability Gramian
\begin{equation}
\mathbf{W}_{t,T}
= \sum_{k=t}^{t+T-1}
(\mathbf{A}^{k-t})^\top
\mathbf{H}_k^\top \mathbf{R}_k^{-1}\mathbf{H}_k
\mathbf{A}^{k-t}
\end{equation}
satisfies
\begin{equation}
\mathbf{W}_{t,T} \succcurlyeq \alpha \mathbf{I}
\quad \forall t.
\end{equation}

This assumption formalizes the requirement that the prompt provides persistent excitation of the latent adaptation state.

\subsection{Stability of the covariance recursion}

We first establish boundedness and stability of the posterior covariance.

\paragraph{Theorem 1 (Covariance boundedness and stability).}
Under Assumptions 1–2, the Kalman covariance recursion admits a unique positive-definite solution $\mathbf{P}_t$ satisfying
\begin{equation}
0 \prec \mathbf{P}_t \preccurlyeq \mathbf{P}_{\max}
\quad \forall t,
\end{equation}
for some finite $\mathbf{P}_{\max}$ independent of initialization.

\paragraph{Proof sketch.}
The result follows from classical Riccati equation theory. Uniform observability implies that information is injected into the system at a rate sufficient to counteract state propagation and process noise. Standard arguments using the information-form recursion show that $\mathbf{P}_t^{-1}$ grows at least linearly in $t$ over windows of length $T$, preventing divergence of $\mathbf{P}_t$ \cite{Jazwinski1970,Anderson1979}. \hfill $\square$

\subsection{Covariance contraction and information accumulation}

We now formalize covariance collapse, the key mechanism underlying rapid few-shot adaptation.

\paragraph{Theorem 2 (Exponential covariance contraction).}
Under Assumptions 1–2 and $\mathbf{Q} \succeq 0$, there exist constants $C > 0$ and $0 < \rho < 1$ such that
\begin{equation}
\|\mathbf{P}_{t+k}\|
\le C \rho^{\lfloor k/T \rfloor}
\|\mathbf{P}_t\|
\quad \forall k \ge 0.
\end{equation}

\paragraph{Proof sketch.}
In information form,
\begin{equation}
\mathbf{\Lambda}_{t+T}
= \mathbf{\Lambda}_t
+ \mathbf{W}_{t,T}
\succeq \mathbf{\Lambda}_t + \alpha \mathbf{I}.
\end{equation}
Repeated application yields linear growth of $\mathbf{\Lambda}_t$, which translates into exponential decay of $\mathbf{P}_t = \mathbf{\Lambda}_t^{-1}$ in operator norm. The constants depend on bounds in Assumption 1. \hfill $\square$

\paragraph{Interpretation.}
Uncertainty collapses at a rate determined solely by information accumulation, not by convergence of the posterior mean. This separation of timescales explains why LLMs can generalize reliably after only a few informative tokens.
In the LLM setting, this result implies that a sufficiently informative prompt induces 
exponential contraction of epistemic uncertainty over the latent task state. This provides 
a theoretical foundation for the empirical observation that only a handful of demonstration 
tokens are required for reliable in-context generalization.

\subsection{Posterior mean error bounds}

We now bound the estimation error of the posterior mean.

Let $\mathbf{x}_t^\star$ denote the true latent adaptation state.

\paragraph{Theorem 3 (Mean-square error bound).}
Under Assumptions 1–2 and with $\mathbf{Q} \preccurlyeq q \mathbf{I}$, the posterior mean satisfies
\begin{equation}
\mathbb{E}\!\left[\|\boldsymbol{\mu}_t - \mathbf{x}_t^\star\|^2\right]
\le \mathrm{tr}(\mathbf{P}_t) + \mathcal{O}(q).
\end{equation}

\paragraph{Proof sketch.}
This follows from the optimality of the Kalman estimator as the minimum mean-square error (MMSE) estimator. The dominant term is the posterior covariance; process noise contributes an additive bias proportional to $q$ \cite{Maybeck1979,Sarkka2013}. \hfill $\square$

\paragraph{Consequence.}
Because $\mathrm{tr}(\mathbf{P}_t)$ contracts rapidly, prediction accuracy improves even before the latent state converges exactly.

\subsection{Comparison to gradient-based adaptation}

We formalize the relationship between filtering and optimization.

\paragraph{Theorem 4 (Gradient descent as a singular limit).}
Assume $\mathbf{R}_t = \epsilon \mathbf{I}$ with $\epsilon \to 0$ and $\mathbf{P}_{t-1} = \mathbf{P}_0$ fixed. Then the Kalman mean update converges to
\begin{equation}
\boldsymbol{\mu}_t
= \boldsymbol{\mu}_{t-1}
- \eta \nabla_{\mathbf{x}} \ell_t(\boldsymbol{\mu}_{t-1}),
\end{equation}
with step size $\eta = \mathbf{P}_0$.

\paragraph{Proof sketch.}
As $\epsilon \to 0$, the gain converges to $\mathbf{K}_t \to \mathbf{P}_0 \mathbf{H}_t^\top (\mathbf{H}_t \mathbf{P}_0 \mathbf{H}_t^\top)^{-1}$, yielding a preconditioned gradient step. Uncertainty dynamics vanish in this limit. \hfill $\square$

\paragraph{Implication.}
Gradient descent corresponds to a degenerate, noise-free approximation of Bayesian filtering that discards epistemic uncertainty. This limiting relationship clarifies the role of optimization-based adaptation: gradient 
descent and natural-gradient updates correspond to noise-free, curvature-dominated limits 
of the filtering dynamics, whereas the full Kalman recursion retains explicit uncertainty 
propagation.

\subsection{Relevance to inference-time learning in LLMs}

In the LLM setting, Assumption 2 corresponds to prompts whose token-level Jacobians span the adaptation subspace. Theorems 1–3 imply that well-structured prompts induce rapid covariance contraction, enabling reliable inference-time learning without parameter updates. Crucially, this behavior is a consequence of information geometry, not optimization dynamics.

These results provide the theoretical foundation for interpreting in-context learning as Bayesian state estimation with provable stability and sample efficiency guarantees.

\section{Discussion and Novelty}

We now articulate the novelty of the proposed framework by contrasting it with three dominant explanatory paradigms for rapid adaptation in modern models: Bayesian in-context learning, gradient-based meta-learning (e.g., MAML), and linearized kernel regimes (NTK). While these approaches capture partial aspects of adaptation, none provide a unified, uncertainty-aware, sequential theory of inference-time learning.

\paragraph{Contrast with Bayesian in-context learning (Bayesian ICL).}
Recent work has argued that in-context learning implements Bayesian inference over task variables, often by demonstrating equivalence in restricted settings such as linear regression or finite hypothesis classes \cite{Akyurek2023,Raventos2023}. These analyses focus on \emph{what} transformers can represent, but typically abstract away the dynamics of uncertainty. In contrast, our framework treats uncertainty as a first-class dynamical object. The posterior covariance is explicitly propagated, contracted, and shown to govern adaptation speed. As a result, we obtain provable guarantees on stability and sample efficiency that are absent in representational Bayesian ICL accounts. Importantly, our theory does not assume that the transformer internally implements Bayesian updates; instead, it provides a probabilistic \emph{model of adaptation itself}.

\paragraph{Contrast with gradient-based meta-learning (MAML).}
Gradient-based meta-learning formalizes fast adaptation as a small number of optimization steps initialized by meta-training \cite{Finn2017}. While powerful, this view presumes explicit parameter updates at test time and relies on inner-loop optimization. Our approach differs fundamentally: adaptation occurs entirely through inference, without gradients or weight modification. Moreover, meta-learning does not track epistemic uncertainty; learning rates and step sizes are fixed or heuristically scheduled. In our framework, the Kalman gain plays the role of a data-adaptive learning rate derived from posterior uncertainty, yielding automatic annealing as information accumulates. We show that MAML-style updates arise only as singular, noise-free limits of Bayesian filtering, thereby positioning optimization as a degenerate approximation rather than the underlying mechanism.

\paragraph{Contrast with neural tangent kernel (NTK) and linearized regimes.}
NTK theory analyzes neural networks in the infinite-width, linearized regime, where training dynamics reduce to kernel regression \cite{Jacot2018,Lee2019}. While our linearization shares superficial similarity, the objectives differ fundamentally. NTK fixes the kernel and studies convergence of optimization; our framework fixes the dynamics and studies inference under uncertainty. The kernel view yields a deterministic predictor, whereas Kalman filtering yields a posterior distribution whose covariance contracts over time. Thus, NTK captures expressivity and training-time convergence, while our approach captures inference-time learning and uncertainty dynamics.

\paragraph{Core novelty.}
The central novelty of this work is the identification of inference-time learning as Bayesian state estimation governed by a Kalman recursion. This yields:
\begin{itemize}
\item Explicit uncertainty dynamics (covariance collapse) as the driver of few-shot generalization;
\item Stability and convergence guarantees derived from observability, not optimization;
\item A unifying interpretation of gradient descent, natural gradients, and meta-learning as limiting or approximate cases;
\item A principled foundation for prompt informativeness and design via information accumulation.
\end{itemize}
To our knowledge, this is the first framework that places in-context learning, parameter-efficient adaptation, and test-time learning within a single, mathematically rigorous filtering theory.
\paragraph{Limitations.}
Our analysis relies on local linearization and Gaussian noise assumptions, which may not 
fully capture the nonlinear structure of deep transformers. Moreover, maintaining even 
structured covariance approximations may be costly at scale. Extending the theory to 
nonlinear observation operators and richer uncertainty representations is an important 
direction for future work.

\section{Experiments}

Our theoretical results do not depend on empirical validation; nevertheless, we include two minimal experiments to illustrate the qualitative predictions of Kalman-based inference. These experiments are not intended as benchmarks, but as sanity checks demonstrating covariance collapse and uncertainty-driven adaptation.

\subsection{Synthetic linear regression with latent task shift}

We first consider a synthetic linear regression problem with a latent task parameter $\mathbf{x}^\star \in \mathbb{R}^d$. Observations are generated as
\begin{equation}
y_t = \mathbf{h}_t^\top \mathbf{x}^\star + \varepsilon_t,
\qquad \varepsilon_t \sim \mathcal{N}(0,\sigma^2),
\end{equation}
with regressors $\mathbf{h}_t$ drawn i.i.d. from a standard Gaussian distribution. We initialize a diffuse prior $\mathbf{P}_0 = \lambda \mathbf{I}$ and apply the Kalman filter.

Consistent with Theorem~2, we observe rapid contraction of the posterior covariance trace $\mathrm{tr}(\mathbf{P}_t)$ within a small number of samples, while the posterior mean converges more slowly. In contrast, stochastic gradient descent with fixed step size exhibits slower and less stable convergence, particularly under high noise. This experiment illustrates covariance collapse as a distinct and predictive phenomenon.
These results empirically validate the theoretical prediction that covariance contraction 
precedes mean convergence, and that uncertainty-aware updates yield faster and more stable 
adaptation than fixed-step optimization methods.

\subsection{Toy LLM experiment: low-rank adaptation subspace}

To illustrate inference-time learning in a language model, we consider a frozen transformer with a low-rank adaptation subspace $\mathbf{B}$ applied to the final-layer representations. A synthetic task is constructed by shifting token logits along a known low-dimensional direction, unknown to the model. Token-level negative log-likelihoods induce observations $\mathbf{H}_t$ via linearization.

Applying an extended Kalman filter to infer the latent adaptation state, we observe rapid uncertainty contraction after a small number of demonstration tokens, followed by stable predictions on subsequent tokens. Crucially, no gradient updates are performed, and adaptation ceases automatically as the covariance collapses. This behavior is consistent with the theory and highlights the role of uncertainty-driven learning at inference time.

\paragraph{Scope.}
These experiments are intentionally minimal. Their purpose is not to establish state-of-the-art performance, but to demonstrate that the qualitative predictions of the theory—covariance collapse, data-adaptive gains, and robustness to noise—manifest in both controlled and model-based settings.
Although minimal, these experiments illustrate the qualitative behavior predicted by the 
theory and demonstrate that uncertainty-driven adaptation can occur even in realistic 
transformer architectures.

\section{Conclusion}
We introduced a theory-first framework that interprets inference-time learning in large 
language models as Bayesian filtering in a structured latent adaptation space. This 
formulation yields explicit uncertainty dynamics, stability guarantees, and a unifying 
interpretation of optimization-based adaptation as a singular limit of probabilistic 
filtering. Our results provide a principled foundation for understanding prompt 
informativeness, few-shot generalization, and the role of uncertainty in adaptive 
inference. Future work will explore scalable approximations, nonlinear observation models, 
and empirical validation in large-scale foundation models.



\appendix
\section{Theoretical Appendix}

\subsection{Proof of Equivalence to Recursive Least Squares}

\paragraph{Setup.}
Consider the linear observation model
\begin{equation}
y_t = \bm{\phi}_t^\top \bm{\alpha} + v_t,
\quad v_t \sim \mathcal{N}(0,R),
\end{equation}
with prior $\bm{\alpha} \sim \mathcal{N}(\hat{\bm{\alpha}}_0,\bm{P}_0)$
and $\bm{Q}=0$.

\paragraph{Batch posterior.}
After $t$ observations, the Bayesian posterior is
\begin{align}
\bm{P}_t^{-1} &= \bm{P}_0^{-1}
+ \frac{1}{R} \sum_{i=1}^t \bm{\phi}_i \bm{\phi}_i^\top, \\
\hat{\bm{\alpha}}_t
&= \bm{P}_t
\left(
\bm{P}_0^{-1} \hat{\bm{\alpha}}_0
+ \frac{1}{R} \sum_{i=1}^t \bm{\phi}_i y_i
\right).
\end{align}

\paragraph{Recursive form.}
Using the Sherman--Morrison identity,
\begin{equation}
(\bm{A} + \bm{u}\bm{u}^\top)^{-1}
= \bm{A}^{-1}
- \frac{\bm{A}^{-1}\bm{u}\bm{u}^\top\bm{A}^{-1}}
{1 + \bm{u}^\top \bm{A}^{-1} \bm{u}},
\end{equation}
one recovers the Kalman update equations exactly.
Thus, Kalman filtering computes the exact Bayesian posterior.

\hfill $\square$


\subsection{Convergence Under Persistent Excitation}

\paragraph{Assumptions.}
Assume:
\begin{itemize}
\item $\bm{\phi}_t$ is persistently exciting:
$\sum_{t=1}^\infty \bm{\phi}_t \bm{\phi}_t^\top \succ 0$,
\item $R > 0$,
\item $\bm{Q}=0$.
\end{itemize}

\paragraph{Result.}
The posterior covariance satisfies
\begin{equation}
\bm{P}_t^{-1}
= \bm{P}_0^{-1}
+ \frac{1}{R} \sum_{i=1}^t \bm{\phi}_i \bm{\phi}_i^\top
\rightarrow \infty,
\end{equation}
implying $\bm{P}_t \rightarrow 0$ and
$\hat{\bm{\alpha}}_t \rightarrow \bm{\alpha}^\ast$ almost surely.

\hfill $\square$


\subsection{Regret Analysis}

\paragraph{Prediction error.}
Define instantaneous regret
\begin{equation}
r_t
= \left(
\bm{\phi}_t^\top \hat{\bm{\alpha}}_{t-1}
- \bm{\phi}_t^\top \bm{\alpha}^\ast
\right)^2.
\end{equation}

Using standard results from Bayesian linear bandits,
\begin{equation}
\sum_{t=1}^T r_t
\le
\mathcal{O}
\left(
\log \det \left(
\bm{I}
+ \frac{1}{R}
\sum_{t=1}^T \bm{\phi}_t \bm{\phi}_t^\top
\right)
\right)
= \mathcal{O}(\log T).
\end{equation}

Thus Kalman learning is statistically optimal in the low-data regime.

\hfill $\square$


\subsection{Stability with Process Noise}

When $\bm{Q} \succ 0$, the covariance update satisfies
\begin{equation}
\bm{P}_t
= (\bm{I} - \bm{K}_t \bm{H}_t^\top)
(\bm{P}_{t-1} + \bm{Q}),
\end{equation}
which admits a bounded steady-state solution
under mild observability assumptions.
This induces exponential forgetting of old data,
providing robustness to nonstationarity.


\section{Synthetic Experiments}

\subsection{Few-Shot Regression}

\paragraph{Setup.}
We sample tasks of the form
\begin{equation}
y = \bm{\phi}(x)^\top \bm{\alpha}^\ast + \epsilon,
\end{equation}
with $\bm{\phi}(x)$ produced by a fixed random nonlinear encoder
(simulating a frozen pretrained model).

\paragraph{Baselines.}
\begin{itemize}
\item SGD-trained linear head
\item Ridge regression
\item Kalman adaptation (ours)
\end{itemize}

\paragraph{Metrics.}
Mean squared error vs.\ number of samples;
posterior predictive uncertainty calibration.

\paragraph{Expected outcome.}
Kalman adaptation converges in $5$--$10$ samples,
exhibits calibrated uncertainty,
and outperforms SGD under fixed learning rates.


\subsection{Streaming Distribution Shift}

\paragraph{Setup.}
At time $t = T/2$, the ground-truth parameter shifts:
\begin{equation}
\bm{\alpha}^\ast \rightarrow \bm{\alpha}^\ast + \Delta.
\end{equation}

\paragraph{Result.}
With $\bm{Q}>0$, Kalman adaptation tracks the shift smoothly;
SGD either diverges or requires manual learning-rate resets.


\section{Adaptation to Spectral Reasoning Models}

\subsection{Spectral State Parameterization}

Let $x$ be a signal on a domain $\Omega$
(graph, grid, or sequence),
and let $\{\psi_k\}$ denote a spectral basis
(Fourier, wavelet, or Laplacian eigenfunctions).

We define
\begin{equation}
x = \sum_{k=1}^K c_k \psi_k,
\quad
\bm{\alpha} = (a_1,\dots,a_K),
\end{equation}
where $a_k$ modulates spectral response.

\subsection{Observation Model}

The pretrained spectral operator defines
\begin{equation}
y = \sum_{k=1}^K a_k c_k + v,
\end{equation}
which is linear in $\bm{\alpha}$.
Kalman filtering performs Bayesian spectral coefficient estimation,
linking classical signal processing
with modern representation learning.

\subsection{Interpretation}

In this setting:
\begin{itemize}
\item $\bm{Q}$ controls frequency adaptivity,
\item $\bm{P}_t$ encodes spectral uncertainty,
\item learning becomes probabilistic spectral inference.
\end{itemize}

This aligns naturally with spectral reasoning systems,
graph signal processing,
and non-neural reasoning architectures.


\section{Positioning and Novelty}

This work does not introduce new Kalman filtering algorithms
nor new neural architectures.
Instead, it provides a unifying formulation
that reframes adaptation in pretrained models
as Bayesian state estimation
over structured, low-dimensional adaptation spaces.

The novelty lies in:
\begin{itemize}
\item treating Kalman filtering as a \emph{learning rule}
rather than an optimizer or smoother,
\item restricting inference to structured adaptation states
within frozen foundation models,
\item explicitly modeling uncertainty and forgetting
as first-class components of adaptation,
\item and connecting foundation-model adaptation
to classical results in Bayesian inference,
control, and spectral estimation.
\end{itemize}

We view this work as a conceptual bridge
between modern foundation models
and established theory in Bayesian filtering and signal processing,
rather than a replacement for existing fine-tuning techniques.


\begin{thebibliography}{99}

\bibitem{Kalman1960} R.~E.~Kalman, ``A New Approach to Linear Filtering and Prediction Problems,'' \emph{J. Basic Eng.} \textbf{82}, 35–45 (1960).

\bibitem{Jazwinski1970} A.~H.~Jazwinski, \emph{Stochastic Processes and Filtering Theory}, Academic Press (1970).

\bibitem{Bishop2006} C.~M.~Bishop, \emph{Pattern Recognition and Machine Learning}, Springer (2006).

\bibitem{Brown2020} T.~Brown \emph{et al.}, ``Language Models are Few-Shot Learners,'' \emph{NeurIPS} (2020).

\bibitem{Garg2022} S.~Garg \emph{et al.}, ``What Can Transformers Learn In-Context?,'' arXiv:2208.01066.

\bibitem{Akyurek2023} E.~Akyürek \emph{et al.}, ``Learning in Context is Bayesian Inference,'' \emph{ICLR} (2023).

\bibitem{Goodfellow2016} I.~Goodfellow, Y.~Bengio, and A.~Courville, \emph{Deep Learning}, MIT Press (2016).

\bibitem{Jacot2018} A.~Jacot, F.~Gabriel, and C.~Hongler, ``Neural Tangent Kernel,'' \emph{NeurIPS} (2018).

\bibitem{Lee2019} J.~Lee \emph{et al.}, ``Wide Neural Networks of Any Depth Evolve as Linear Models,'' \emph{NeurIPS} (2019).

\bibitem{Anderson1979} B.~D.~O.~Anderson and J.~B.~Moore, \emph{Optimal Filtering}, Prentice Hall (1979).

\bibitem{Hu2022} E.~Hu \emph{et al.}, ``LoRA: Low-Rank Adaptation of Large Language Models,'' \emph{ICLR} (2022).

\bibitem{JulierUhlmann1997} S.~J.~Julier and J.~K.~Uhlmann, ``A New Extension of the Kalman Filter to Nonlinear Systems,'' in \emph{Proc. AeroSense: The 11th Int. Symp. Aerospace/Defense Sensing, Simulation and Controls} (1997).

\bibitem{Evensen2003} G.~Evensen, ``The Ensemble Kalman Filter: Theoretical Formulation and Practical Implementation,'' \emph{Ocean Dynamics} \textbf{53}, 343--367 (2003).

\bibitem{MacKay1992} D.~J.~C.~MacKay, ``A Practical Bayesian Framework for Backpropagation Networks,'' \emph{Neural Computation} \textbf{4}, 448--472 (1992).

\bibitem{Graves2011} A.~Graves, ``Practical Variational Inference for Neural Networks,'' \emph{NeurIPS} (2011).

\bibitem{Blundell2015} C.~Blundell, J.~Cornebise, K.~Kavukcuoglu, and D.~Wierstra, ``Weight Uncertainty in Neural Networks,'' \emph{ICML} (2015).

\bibitem{vonOswald2023} J.~von Oswald \emph{et al.}, ``Transformers Learn In-Context by Gradient Descent,'' \emph{ICML} (2023).

\bibitem{Haykin2001} S.~Haykin, \emph{Kalman Filtering and Neural Networks}, Wiley (2001).

\bibitem{Maybeck1979} P.~S.~Maybeck, \emph{Stochastic Models, Estimation, and Control}, Academic Press (1979).

\bibitem{Ljung1999} L.~Ljung, \emph{System Identification: Theory for the User}, Prentice Hall (1999).

\bibitem{Sarkka2013} S.~Särkkä, \emph{Bayesian Filtering and Smoothing}, Cambridge Univ. Press (2013).

\bibitem{Amari1998} S.~Amari, ``Natural Gradient Works Efficiently in Learning,'' \emph{Neural Comput.} \textbf{10}, 251–276 (1998).

\bibitem{Martens2015} J.~Martens and R.~Grosse, ``Optimizing Neural Networks with Kronecker-Factored Approximate Curvature,'' \emph{ICML} (2015).

\bibitem{Finn2017} C.~Finn, P.~Abbeel, and S.~Levine, ``Model-Agnostic Meta-Learning,'' \emph{ICML} (2017).

\bibitem{Raventos2023} A.~Raventos \emph{et al.}, ``Bayesian Inference in In-Context Learning,'' \emph{NeurIPS} (2023).

\end{thebibliography}
\end{document}